\documentclass[journal]{IEEEtran}
\usepackage[pdftex]{graphicx}
\usepackage[colorlinks,linkcolor=blue]{hyperref}
\usepackage{amsmath}
\usepackage{amsfonts,amssymb}
\usepackage{algorithmic}
\usepackage{multirow}
\usepackage{graphicx}
\usepackage{cite}
\usepackage{flushend}
\ifCLASSOPTIONcompsoc
\else
\fi
\begin{document}
%
\title{Scale-Aware Network with Regional and Semantic Attentions for Crowd Counting under Cluttered Background}
%
%
%

\author{Qiaosi Yi\textsuperscript{$\dagger$}, Yunxing Liu\textsuperscript{$\dagger$}, Aiwen Jiang\textsuperscript{*}, Juncheng Li, Kangfu Mei, Mingwen Wang
\thanks{$\dagger$: Equal contribution. *: The corresponding author.}
\thanks{Q. Yi, and J. Li are with the Shanghai Key Laboratory of Multidimensional Information Processing, East China Normal University, Shanghai, China, and also with the school of Computer Science and Technology, East China Normal University, Shanghai, China. (E-mail: qiaosiyijoyies@gmail.com, cvjunchengli@gmail.com)}
\thanks{Y. Liu, A. Jiang, and M. Wang are with the School of Computer and Information Engineering, Jiangxi Normal University, Nanchang, China. (E-mail: 201841600023@jxnu.edu.cn, jiangaiwen@jxnu.edu.cnm, mwwang@jxnu.edu.cn)}
\thanks{K. Mei is with the Department of Mathematics, The Chinese University of
Hong Kong (ShengZheng), ShengZheng, China. (E-mail: kangfumei@link.cuhk.edu.cn)}
}

%
%

\markboth{Journal of \LaTeX\ Class Files,~Vol.~14, No.~8, August~2015}%
{Shell \MakeLowercase{\textit{et al.}}: Bare Demo of IEEEtran.cls for IEEE Journals}
%



\maketitle

\begin{abstract}
Crowd counting is an important task that shown great application value in public safety-related fields, which has attracted increasing attention in recent years. 
In the current research, the accuracy of counting numbers and crowd density estimation are the main concerns.
Although the emergence of deep learning has greatly promoted the development of this field, crowd counting under cluttered background is still a serious challenge.
In order to solve this problem, we propose a Scale-Aware Crowd Counting Network (SACCN) with regional and semantic attentions. 
The proposed SACCN distinguishes crowd and background by applying regional and semantic self-attention mechanisms on the shallow layers and deep layers, respectively.
Moreover, the asymmetric multi-scale module (AMM) is proposed to deal with the problem of scale diversity, and regional attention based dense connections and skip connections are designed to alleviate the variations on crowd scales.
Extensive experimental results on multiple public benchmarks demonstrate that our proposed SACCN achieves satisfied superior performances and outperform most state-of-the-art methods. 
All codes and pre-trained models will be released soon.
\end{abstract}


%
\IEEEpeerreviewmaketitle

\section{Introduction}
High-density crowds are prone to public accidents and accelerate the spread of diseases.
In the past year, the COVID-19 has swept the world. In order to prevent its spread again, it is of great significance to monitor and analyze high-density crowds and provide timely and effective evacuation.
As one of the most important assistance for the analysis of high-density population, crowd counting is conducive to real-time guidance and control. Therefore, it has a wide range of applications in public safety-related fields and has attracted widespread attention in the field of computer vision.
The task of crowd counting is aims to estimate the number and the distribution of people in a crowd. 
In the past, the mainstream methods relied on object detection~\cite{dollar2011pedestrian, viola2004robust, chen2012feature} and regression~\cite{boominathan2016crowdnet, chen2013cumulative}. 
Recently, with the success of deep learning, crowd counting tasks have also made significant progress~\cite{zhang2015cross, zhang2016single, li2018csrnet, zheng2018cross}. 
However, it is still a challenging task since crowd scale variation and cluttered backgrounds is very common in the real world.
\begin{figure}[!t]
	\centering
	\includegraphics[scale=0.4]{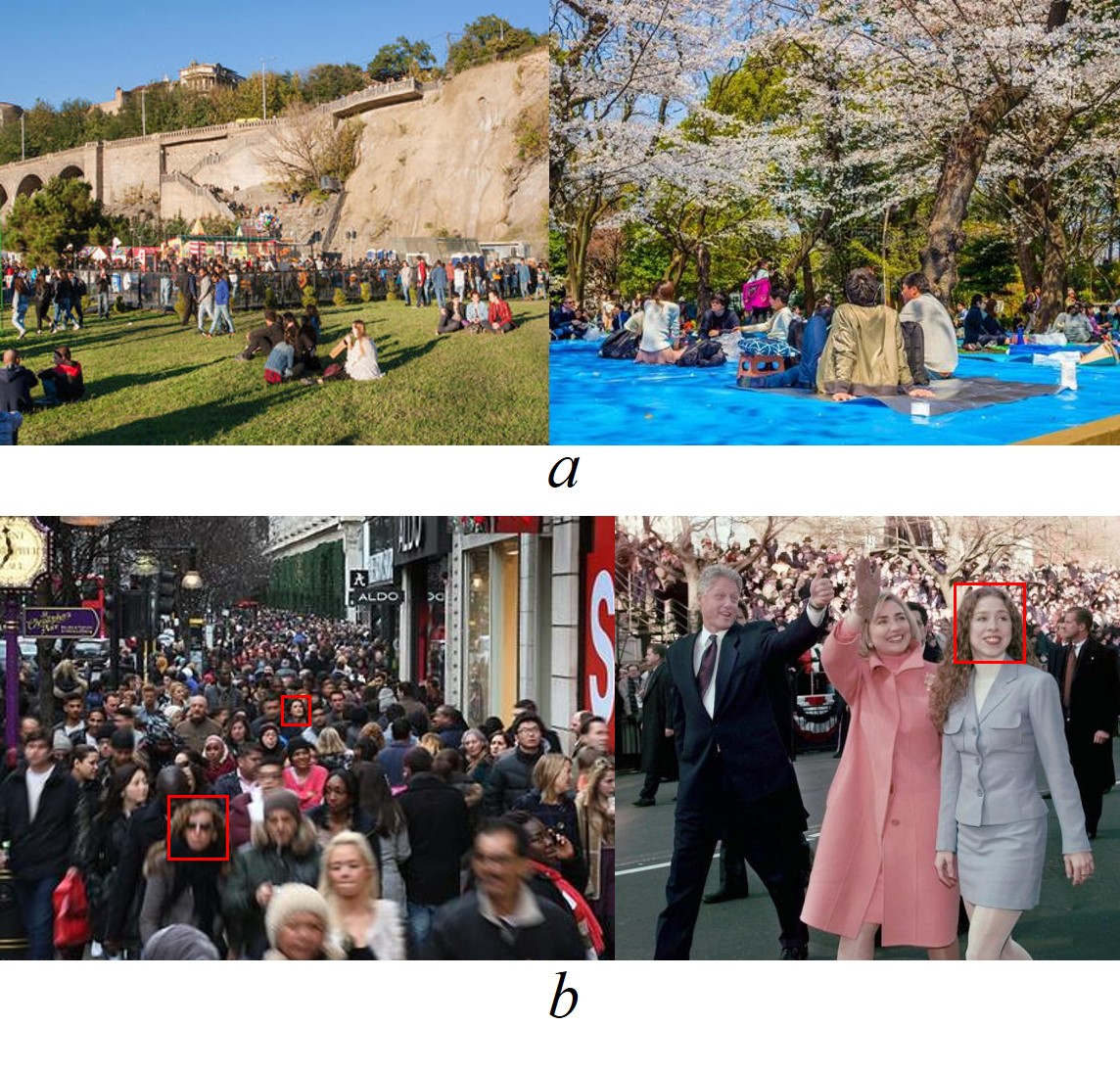} 
	\caption{Two challenges on crowd counting task. a) Cluttered backgrounds. b) Variation on crowd scale.}
	\label{fig1}
\end{figure}

As shown in Fig.~\ref{fig1} (a), the background of the crowd scene is very complicated. 
Due to the gathering and interference of objects other than people, certain parts of the crowd are not easy to be distinguished from a cluttered background. 
In this case, crowd counting methods are prone to fail accurately counting the crowd and evaluating its distribution. 
As shown in Fig.~\ref{fig1} (b), the sizes of persons in the crowd have large variations, either caused by camera perspective or camera position. 
All these unexpected variations put forward rigorous requirements on the robustness of crowd counting algorithms. 
Therefore, in this paper, we aim to design effective solutions for these two obstacles.

To alleviate the counting errors caused by cluttered backgrounds, current mainstream solutions~\cite{liu2018decidenet, sindagi2019ha, liu2019adcrowdnet, miao2020shallow} adopt visual attention mechanism. 
In the latest works, such as SDANet~\cite{miao2020shallow} and HA-CCN~\cite{sindagi2019ha}, researchers found that background and crowd were easier to be distinguished on shallow feature maps. 
Therefore, they suggested to apply pixel-wise attention on low-level feature maps. 
However, as we know, features in different channels contribute different discriminative information on classifying cluttered backgrounds and crowds. 
Limitedly applying pixel-attention is not enough. In this paper, we proposed to design region attention module (RAM) on shallow layers, in which channel-wise attention and pixel-attention collaborate to highlight crowd regions. 

Moreover, we argue that distinguishing clutter backgrounds and crowd foreground can be taken as a semantic classification task. According to ~\cite{lin2017feature}, features in deep layers have richer semantic information and less noise than shallow-layer features for semantic categorization. Therefore, we further propose a semantic attention module (SAM) on deep layers, in which spatial self-attention and channel-wise self-attention collaborate to identify persons in semantic space.

To alleviate the negative influences from crowd scale variations, current mainstream works proposed to utilize strategies like multi-column CNN~\cite{zhang2016single}, stacked multi-branch blocks~\cite{sindagi2017cnn, shi2019revisiting} and multi-scale feature fusion~\cite{sindagi2019multi, miao2020shallow}. However, the crowd scale variation is still far from being solved. As we know, features of different resolutions at different levels redundantly contain complementary information for crowd identification. Therefore, in this paper, we proposed to construct our crowd counting network in a U-like encoder-decoder structure for adequate feature reuse. Compared with traditional U-like network, the big differences are that, in the encoder stage, regional attention based dense connections are constructed to strength information fusion specifically for crowd identification. Regional attention based skip-connections between encoder and decoder stages and asymmetric multi-scale residual blocks in the decoder stage are originally proposed to account for information aggregation of different resolutions.   

The main contributions can be summarized as followings:
\begin{itemize}
	\item We propose a Scale-Aware Crowd Counting Network (SACCN), which is concentrated on overcoming adverse influences from cluttered background and scale variations in crowd scenes. Comprehensive experiments demonstrate that our SACCN achieves superior performance.
	\item We design two attention modules for crowd identification under cluttered backgrounds. One is the regional attention module (RAM), which calculates spatial and channel-wise attention on shallow features. The other is the semantic attention module (SAM) that imposes both spatial and channel self-attention on deep feature maps. These two attention modules are collaboratively employed in cascade and can effectively highlight crowd areas from cluttered backgrounds.
	\item We propose an asymmetric multi-scale module (AMM) to deal with the problem of scale variations. The module employs asymmetric convolutions with different scales to learn crowd density maps. Ablation experiments confirmed that it can effectively improve the model performance and reduce the model parameters.
	\item We design the regional attention based connections for complementary feature reuse, including the dense connections among shallow layers of the encoder and the skip connections between encoder and decoder. Ablation studies demonstrate that both connections play important roles in performance improvement.
\end{itemize}

The remainders are organized as followings. Related works on crowd counting are briefly reviewed in Section~\ref{relatedwork}. Implementation details of the proposed methods are described in Section~\ref{implementations}. Experimental results and analysis are discussed in Section~\ref{experiments}. Conclusion is presented in Section~\ref{conclusion}.

\section{Related Work}
\label{relatedwork}
The methods of crowd counting can be roughly divided into three paradigms: \textit{traditional detection-based methods}, \textit{regression-based methods}, and \textit{modern CNN-based methods}. In this section, we briefly review some typical related works for completeness. 

\subsection{Detection-based Methods}
Traditional detection-based methods~\cite{leibe2005pedestrian, dollar2011pedestrian} mainly focused on the characteristics of people, like face, body, etc. Methods of this paradigm usually employed sliding window-based detection schemes. Hand-crafted features like Haar wavelets~\cite{viola2004robust} and Histogram of Gradient (HoG)~\cite{dalal2005histograms} were extracted from images for object classification. Generally, these methods can provide accurate count in scenes with sparse objects, but in scenes with complex backgrounds and dense objects. These methods cannot work well when there exist severe crowd occlusion and large variations on crowd scales. 

\subsection{Regression-based Methods}

In order to deal with complex scenes with dense crowds, some methods~\cite{chan2009bayesian, chen2012feature} proposed a regression method to avoid hard detection. These methods generally learned regression functions from local features of image patches to counting numbers. In these cases, spatial distribution information of people was often ignored.

Some other researchers~\cite{lempitsky2010learning, pham2015count} also proposed density-based regression methods to take full advantage of annotation. However, variations in the crowd scale and cluttered backgrounds are still the main obstacles.

\begin{figure*}[!t]
	\centering
	\includegraphics[scale=0.45]{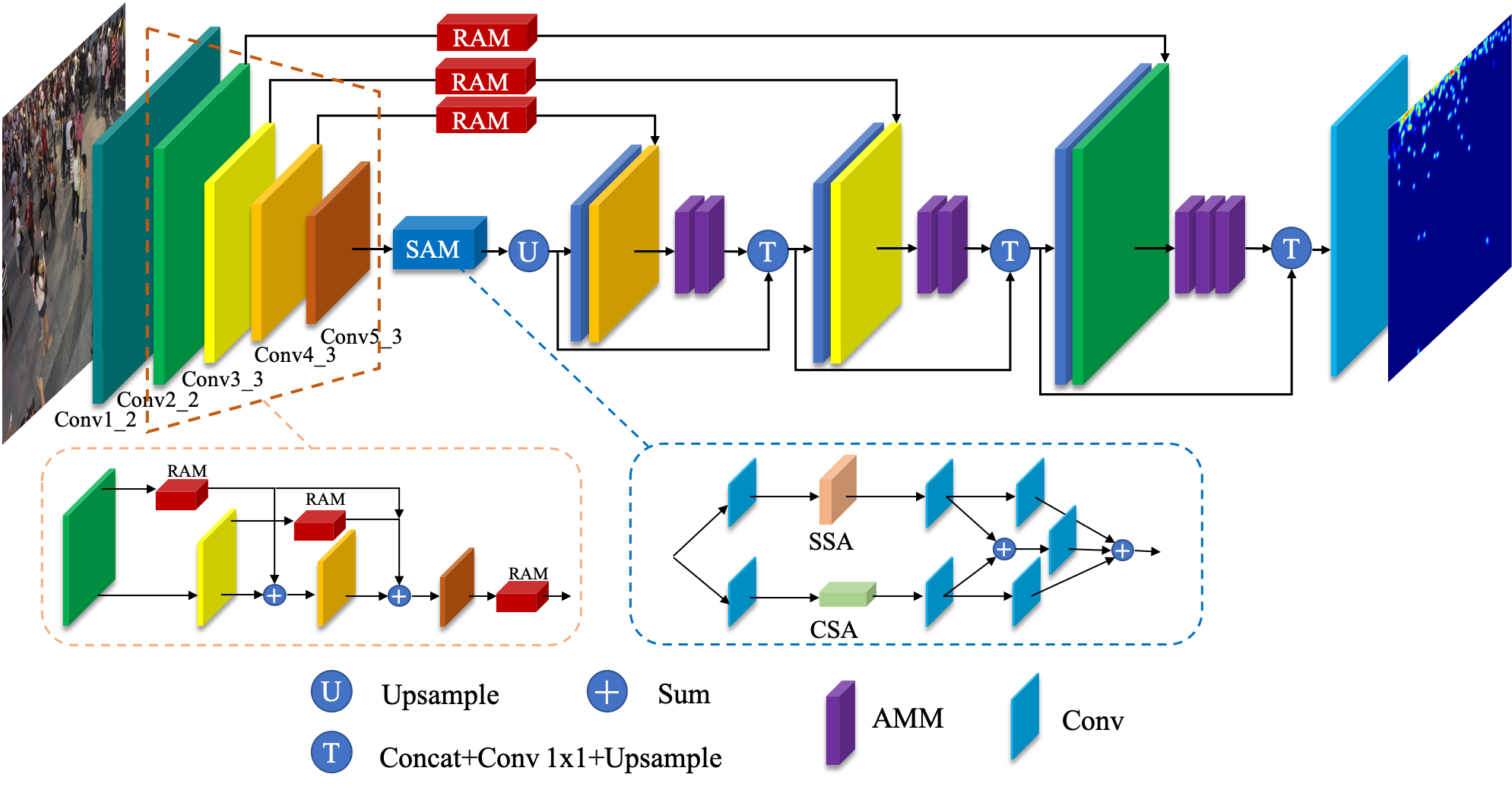} 
	\caption{The complete architecture of our proposed Scale-Aware Crowd Counting Network (SACCN).}
	\label{fig2}
\end{figure*}

\subsection{CNN-based Methods}

With the success of the convolution neural network (CNN) in the computer vision field, the crowd counting community has also realized to make use of the strong representation learning ability of CNN to estimate crowd density. CNN-based methods~\cite{zhang2016single,li2018csrnet,miao2020shallow,sindagi2019ha} have shown significant improvements over previous methods. 

Most of CNN-based methods address the problems of crowd scale variation and cluttered backgrounds through strategies such as multi-column CNN~\cite{zhang2016single}, stacked multi-branch blocks~\cite{sindagi2017cnn}, multi-scale feature fusion~\cite{sindagi2019multi, miao2020shallow}, selective regression~\cite{sam2017switching,sindagi2017generating} and visual attention~\cite{liu2019adcrowdnet}.

Crowdnet~\cite{boominathan2016crowdnet} was the first work to generate a density map by using the convolution network. It adopted a dual-column structure to solve crowd scale variation problems. MCNN~\cite{zhang2016single} adopted a multi-column structure for crowd counting. CSRNet~\cite{li2018csrnet} adopted dilated convolutions that expanded the receptive field without parameters increasing. SANet~\cite{cao2018scale} stacked multi-branch blocks as the feature extractor to overcome variations on crowd scales. MBTTBF-SCFB\cite{sindagi2019multi} propose a multilevel bottom-top and top-bottom fusion method. SDANet~\cite{miao2020shallow} found that using channel-wise attention in shallow layers can distinguish background and crowd better than in deep layers. HA-CCN~\cite{sindagi2019ha} uses spatial attention in different levels of VGGNet~\cite{simonyan2014very} to distinguish the background and the crowd.

The closest methods to ours are the latest SDANet~\cite{miao2020shallow} and the HA-CCN~\cite{sindagi2019ha}. However, there exist many big differences. For example, they only apply single attention, either limited channel-wise attention or spatial attention. In contrast, our method combines channel-wise attention and spatial attention together on feature maps at shallow layers to distinguish complicated backgrounds and crowd. Moreover, we employ spatial and channel-wise self-attention on deep layers at the same time to enhance discriminability of semantic representations for crowd.

\section{Scale-Aware Crowd Counting Network}
\label{implementations}
In this section, we describe the proposed Scale-Aware Crowd Counting Network(SACCN) in detail. As shown in Fig.~\ref{fig2}, our method is mainly composed of three parts: \textit{encoder part}, \textit{decoder part} and \textit{regional attention based connections}. Specifically, in the encoder part, the pre-trained VGGNet16\cite{simonyan2014very} model is employed as the backbone network for feature extraction. Regional attention based dense connections are constructed to specifically strengthen information fusion for crowd identification. In the decoder part, asymmetric multi-scale residual blocks are employed to account for information aggregation of different resolutions. Regional attention based skip-connections between encoder and decoder layers are for feature reuse and crowd highlights. More details are explained in the following subsections.

\subsection{Attention Modules}
Generally, crowd scene images not only contain people, but also backgrounds such as buildings, trees, cars, etc. Attention has been proved to be a very helpful mechanism for crowd counting~\cite{miao2020shallow,sindagi2019ha}, especially when dealing with clutter backgrounds. Therefore, we specifically propose both regional attention and semantic attention modules for crowd representations highlighting.

\subsubsection{Regional Attention Module} 
\begin{figure}[!t]
	\centering
	\includegraphics[scale=0.25]{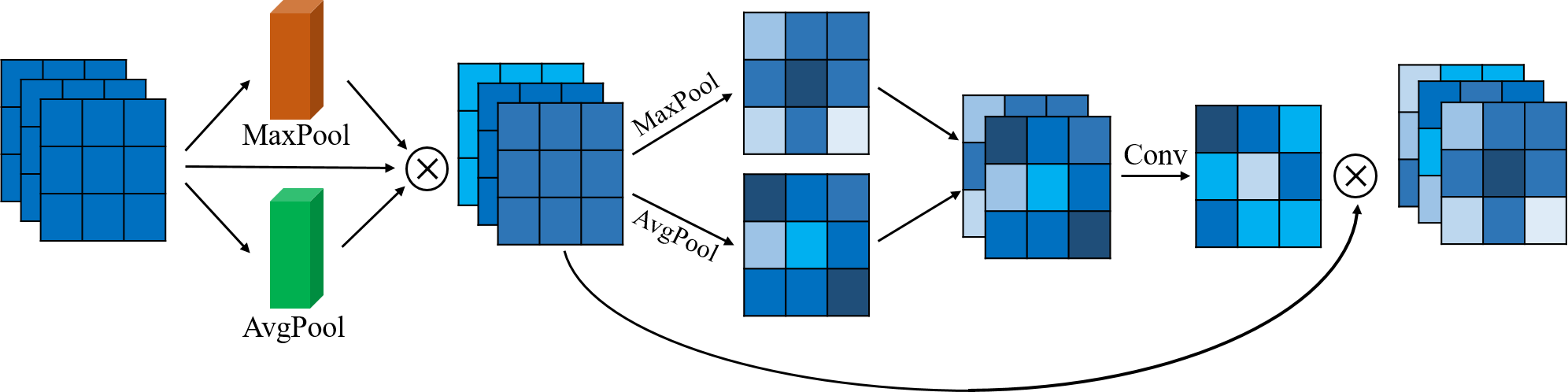} 
	\caption{The structure of Regional Attention Module (RAM).}
	\label{ram}
\end{figure}

The structure of regional attention module (RAM) is shown in Fig.~\ref{ram}. It cascade combines channel-wise attention and spatial attention in a holistic block. The channel-wise attention learns to capture the importance of different feature channels, helping suppress unnecessary noisy information. Spatial attention is to focus on crowd relevant areas. Through explicitly employing spatial attention on the feature maps processed by channel-wise attention, the proposed network can be more easily encouraged to focus on representations learning for the interested crowd areas.

Formally, given a feature map $X \in \mathbb{R}^{C\times H \times W}$, in channel-wise attention, average-pooling and max-pooling are firstly performed to produce pooled features $Y_{avg} \in \mathbb{R}^{C\times 1 \times 1}, Y_{max} \in \mathbb{R}^{C\times 1 \times 1}$ respectively. Then, the $Y_{avg}$ and  $Y_{max}$ pass through two fully-connected (FC) layers to produce $F_{avg}$ and $F_{max}$, as shown Equation~\ref{eq1}. Parameters of the two $\text{FC}$ layers are shared across the two paths.
\begin{equation}
\small
	\begin{aligned}
		F_{avg} &= \text{FC}_{2}(\text{ReLU}(\text{FC}_{1}(Y_{avg}))), \\
		F_{max} &= \text{FC}_{2}(\text{ReLU}(\text{FC}_{1}(Y_{max}))).
		\label{eq1}
	\end{aligned}
\end{equation}
Then, channel-wise attention vector $S_{a}$ is generated according to Equation~\ref{Sa}.
\begin{equation}
\small
	\label{Sa}
	S_{a} = sigmoid(F_{avg}+F_{max}).
\end{equation}
Intermediate feature map $X_{c}$ is therefore obtained through element-wise weighting on input $X$. 
\begin{equation}
\small
	X_{c} = X \otimes S_{a}.
\end{equation}

Similar to channel-wise attention, spatial attention in parallel employs max-pooling and average-pooling on $X_c$ in channel direction to produce $\hat{Y}_{max} \in \mathbb{R}^{1\times H \times W}$ and $\hat{Y}_{avg} \in \mathbb{R}^{1\times H \times W}$ respectively. Then, $\hat{Y}_{avg}$ and $\hat{Y}_{max}$ are concatenated in channel to get $\hat{Y_s}$. We employ a convolution with kernel size $3 \times 3$ on it to produce a spatial attention map $S_{b}$.
\begin{equation}
\small
	S_{b} = sigmoid(conv(\hat{Y_s})).
\end{equation}
Finally, the feature map $X_{s}=X_{c} \otimes S_{b}$ is taken as the output of regional attention module, where $\otimes$ represents elementary multiplication.  

\subsubsection{Semantic Attention Module} 
\begin{figure}[!t]
	\centering
	\includegraphics[scale=0.3]{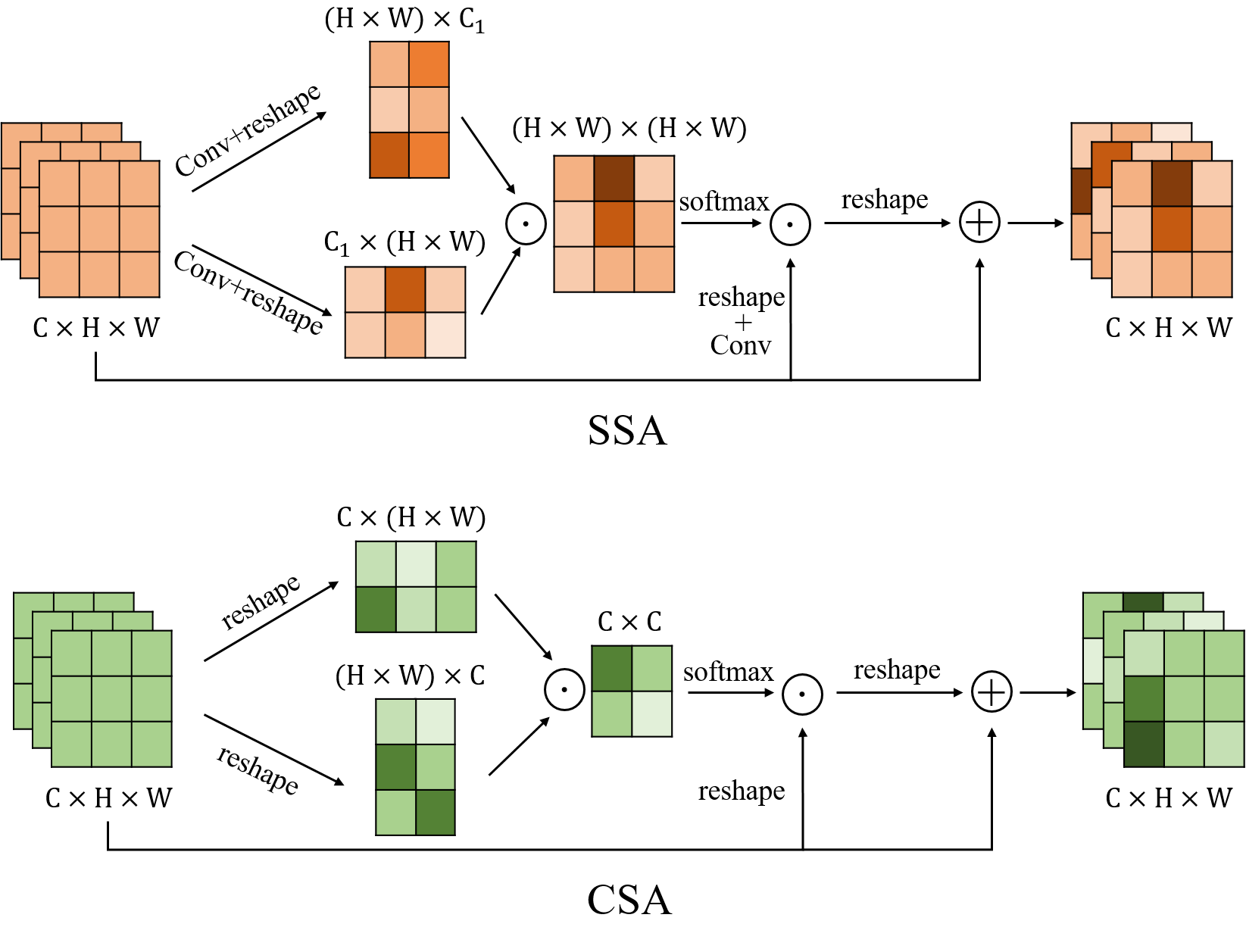} 
	\caption{The structure of the Spatial Self-Attention (SSA) and Channel-wise Self-Attention (CSA). The $\odot$ represents the matrix multiplication.}
	\label{sam}
\end{figure}

Motivated by the success of self-attention~\cite{wang2018non,zhang2019self,fu2019dual} in many high-level vision tasks, we explore to utilize it on crowd counting tasks.

The structure of the proposed semantic attention module (SAM) is shown in Fig.~\ref{fig2}. It is mainly composed of spatial self-attention (SSA), channel-wise self-attention (CSA), and some necessary convolutions. The structural details of SSA and CSA are shown in Fig.~\ref{sam}.  

In case of spatial self-attention (SSA), given a feature map $X \in \mathbb{R}^{C\times H \times W}$, two channel-wise features are firstly transformed to $C_1$-dimension feature space, both through convolution with kernel size $1 \times 1$. Then, the resulted feature map is reshaped to $X_1 \in \mathbb{R}^{(H W)\times C_1}$ and $X_2 \in \mathbb{R}^{C_1 \times (H W)}$. A spatial attention map $W_s$ is consequently generated through feature correlation as shown in Equation~\ref{Ws}. 
\begin{equation}
\small
	\label{Ws}
	W_s = softmax(X_1 \odot X_2),
\end{equation}
where, $\odot$ represents matrix product.

In parallel, we perform feature transformation on input $X$ through a convolution with kernel size $1 \times 1$, and generate an intermediate feature $X_3 \in \mathbb{R}^{C\times (H W)}$. 

The output $X_{ssa}$ of SSA block is then calculated in residual mode, as shown in Equation~\ref{ssa}.
\begin{equation}
\small
	\label{ssa}
	X_{ssa} = X + reshape(X_3 \odot W_s),
\end{equation}
where, $\odot$ represents matrix product. 

In case of channel-wise self-attention (CSA), given a feature map $X \in \mathbb{R}^{C\times H \times W}$, we firstly reshape the feature map to produce $X_4 \in \mathbb{R}^{C\times (H W)}$ and $X_5 \in \mathbb{R}^{(H W)\times C}$. Then, we perform matrix product and softmax operations to generate a channel-wise attention vector $W_c$, as shown in Equation~\ref{wc}.
\begin{equation}
	\label{wc}
	W_c =softmax(X_4 \odot X_5).
\end{equation}

The $X_4 \in \mathbb{R}^{(H W)\times C}$ is weighted by $W_c$. The output $X_{csa}$ of CSA is therefore generated according to Equation~\ref{csa}.
\begin{equation}
\small
	\label{csa}
	X_{csa} = X + reshape(X_4 \odot W_c).
\end{equation}

\subsection{Asymmetric Multi-scale Module}
\begin{figure}[!t]
	\centering
	\includegraphics[scale=0.4]{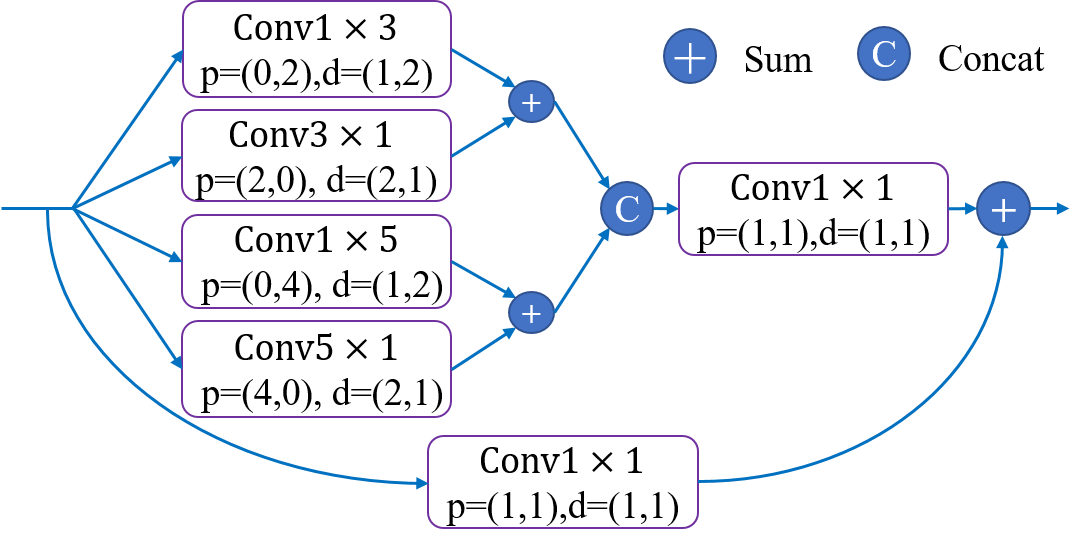} 
	\caption{The structure of the Asymmetric Multi-scale Module (AMM). $Conv k_1\times k_2$ represents an asymmetric convolution with kernel size of $k_1\times k_2$, $p$ is the padding and $d$ stands of the dilated rate.}
	\label{AMM}
\end{figure}

\begin{table*}[h]
	\setlength{\tabcolsep}{5mm}
	\centering
	\caption{Comparison results of different methods on the ShanghaiTech Part\_A/Part\_B, UCF\_CC\_50 and UCF\_QNRF. Best results and Second results are highlighted.}
	\begin{tabular}{|c|c|c|c|c|c|c|c|c|}
	\hline
		\multirow{2}{*}{Method} & \multicolumn{2}{|c|}{Part\_A} & \multicolumn{2}{|c|}{Part\_B} & \multicolumn{2}{|c|}{UCF\_CC\_50} & \multicolumn{2}{|c|}{UCF\_QNRF} \\
		\cline{2-9}
		                        & MAE & MSE & MAE & MSE & MAE & MSE & MAE & MSE\\
		\hline
		MCNN\cite{zhang2016single}(CVPR'16)         &110.2 & 173.2 & 26.4 & 41.3 & 377.6 & 509.1 & 277.0 & 426.0\\
		CP-CNN\cite{sindagi2017generating}(ICCV'17) & 73.6 & 106.4 & 20.1 & 30.1 & 295.8 & 320.9 & -     & -    \\
		Switch-CNN\cite{sam2017switching}(CVPR'17)  & 90.4 & 135.0 & 21.6 & 33.4 & 318.1 & 439.2 & 228.0 & 445.0\\
		CSRNet\cite{li2018csrnet}(CVPR'18)          & 68.2 & 115.0 & 10.6 & 16.0 & 266.1 & 397.5 & 135.4 & 207.4\\
		SANet\cite{cao2018scale}(ECCV'18)           & 67.0 & 104.5 & 8.4  & 13.6 & 258.5 & 334.9 & -     & -    \\
        ic-CNN\cite{ranjan2018iterative}(ECCV'18)   & 69.8 & 117.3 & 10.7 & 16.0 & 260.9 & 365.5 & -     & -    \\
		HA-CCN\cite{sindagi2019ha}(TIP'19)         & 62.9 & \textbf{94.9}  & 8.1  & 13.4 & 256.2 & 348.4 & 118.1 & 180.4\\
		MACC\cite{jiang2019mask}(TCSVT'19)          & 61.8 & 100.0 & 8.6  & 13.3 & 245.4 & 349.3 & -     &-\\
		COBC\cite{liu2019counting}(TCSVT'19)        & 62.8 & 102.0 & 8.6 & 16.4  & 239.6 & 322.2 & 118.0 & 192.0\\
		ADCrowdNet\cite{liu2019adcrowdnet}(CVPR'19)     & 63.2 & 98.9  & 8.2  & 15.7 & 266.4 & 358.0 & -     & -    \\
        CAN\cite{liu2019context}(CVPR'19)        & 61.3 & 100.0 & 7.8  & 12.2 & 212.2 & \textbf{243.7} & 107.0 & 183.0\\
		PSDDN\cite{liu2019point}(CVPR'19)           & 85.4 & 159.2 & 16.1 & 27.9 & 359.4 & 514.8 & - & -\\
	    SPN+L2SM\cite{xu2019learn}(ICCV'19)           & 64.2 & 98.4  & \textbf{7.2}  & 11.1 & \textbf{188.4} & 315.3 & -     & -    \\
        MBTTBF-SCFB\cite{sindagi2019multi}(ICCV'19)      & \textbf{60.2} & \textbf{94.1}  & 8.0  & 15.5 & 233.1 & 300.9 & \textbf{97.5}  &  \textbf{165.2}\\
   		DSSINet\cite{liu2019crowd}(ICCV'19)         & 60.6 & 96.1  & 6.9  & 10.4 & 216.9 & 302.4 & 99.1 & 159.2\\ 
		SDANet\cite{miao2020shallow}(AAAI'20)       & 63.6 & 101.8 & 7.8  & \textbf{10.2} & 227.6 & 316.4 & -     & -    \\
		DUBNet\cite{oh2020crowd}(AAAI'20)           & 64.4 &	106.8 & 7.7  & 12.5 & 243.8 & 329.3 & 105.6 & 180.5\\
		HSRNet\cite{zou2020crowd}(ECAI'20) &  62.3 & 100.3 & 7.2 & 11.8 & - & - & - & -\\
		KDMG \cite{wan2020kernel}(T-PAMI'20)          & 63.8 & 99.2 & 7.8 & 12.7 & - & - & 99.5 & 173\\
		\hline
		SACCN (Ours)                               & \textbf{59.2} & 98.0  & \textbf{6.8} & \textbf{10.5} & \textbf{178.0} & \textbf{258.5} & \textbf{96.1} & \textbf{167.8}\\
		\hline
	\end{tabular}
	\label{table1}
\end{table*}
In crowd counting related works, multi-scale modules are widely used to solve scale variations. Inspired by Inception-v3~\cite{szegedy2016rethinking} and ACNet~\cite{ding2019acnet}, we propose to use asymmetric convolutions for density estimation. As shown in Fig.~\ref{AMM}, we design an asymmetric multi-scale module (AMM) which approximates conventional convolution with a kernel size of $k \times k $ by the summation of two asymmetric convolutions with kernel size of $k\times1$ and $1\times k$. The $1\times 1$ convolution is employed for feature fusion after the concatenation of multi-branch convolutions. 

Moreover, in order to enlarge the receptive field without parameters increase, all asymmetric convolutions are set to be dilated convolutions. In this paper, the dilation rate is set to be 2. To ensure the size of the output feature keeping the same as input, we set the padding for each convolution, such as the padding (0,2) for $Conv1\times3$ is and (0,4) for $Conv1\times5$. 

\subsection{Regional Attention based Connections}
In previous work~\cite{sindagi2019multi, miao2020shallow}, multi-scale fusion is verified to be helpful for crowd counting. Therefore, in order to make full use of the redundant and complementary information from different layers, we construct several kinds of regional attention based connections in our crowd counting network. The connections are shown in Fig.~\ref{fig2}.

\subsubsection{Regional Attention Based Dense Connections} We employ the first 13 layers of VGG16 as the backbone network for feature extraction. Besides, we add extra dense connections among layers $Conv2\_2$, $Conv3\_3$, $Conv4\_3$, and $Conv5\_3$ for multi-resolution feature reuse. In order to highlight crowd information while suppress background, we utilize region attention mechanism in dense connection paths, as shown in Fig.~\ref{fig2}. We formulate the process in Equation~\ref{Eq.dense}.
\begin{equation}
\small
	\begin{aligned}
		I_3 &= Conv2\_2\downarrow_{2},\\
		I_4 &= \texttt{RAM}(Conv2\_2)\downarrow_{4} + Conv3\_3\downarrow_{2},\\
		I_5 &=\texttt{RAM}(Conv2\_2)\downarrow_{8} + \texttt{RAM}(Conv3\_3)\downarrow_{4} + Conv4\_3\downarrow_{2},
	\end{aligned}
	\label{Eq.dense}
\end{equation}
where, $I_k$ is the input of the "$Conv~\textbf{\textit{k}}$" layer in encoder. $\texttt{RAM}$ represents the regional attention module, $\downarrow_{m}$ means down-sampling by $m$ times .

\subsubsection{Regional Attention Based Skip-connections} 
As we know, when the encoder going deeper, the resolutions of the feature map are becoming lower due to the down-sampling process. For adequate feature reuse, skip-connections are adopted in a conventional U-like encoder-decoder network. However, from the viewpoint of information reuse, convention skip connections transfer information uniformly without specific focus. In our method, we propose to introduce regional attention into skip-connections to highlight crowd identification. 

For the convenience of description, we denote the $k^{th}$ layer in the decoder having the same size as its counterpart layer "Conv $k$" in the encoder. We formulate the skip connections process as Equation~\ref{Eq.skipDK} and Equation~\ref{Eq.skipEk}.
\begin{equation}
\small
	D_k = \left\{
	\begin{aligned}
		& Conv_{1\times1}([\texttt{AMM}(E_k);D_{k+1}])\uparrow^{2}, k=4,3,2\\
		&\texttt{SAM}(E_5)\uparrow^{2}, k=5
	\end{aligned}
	\right.
	\label{Eq.skipDK}
\end{equation}

\begin{equation}
\small
	E_k = \left\{
	\begin{aligned}
		&\texttt{RAM}(Conv \texttt{k}) + D_{k+1}, k=4,3,2\\
		&\texttt{RAM}(Conv5), k=5
	\end{aligned}
	\right.
	\label{Eq.skipEk}
\end{equation}
where, $D_k$ stands the output of the $k^{th}$ counterpart layer in decoder, $E_k$ represents the fused features after skip connections at the $k^{th}$ counterpart layer in decoder. RAM (or SAM) represents the Regional Attention Module (or Semantic Attention Module). AMM represents Asymmetric Multi-scale Module. $Conv_{1\times1}$ represents convolution layer with kernel size of $1\times1$. "$[*;*]$" represents concatenation in channel. $\uparrow^k$ represents $k$ times Upsampling.
\section{Experiments}
\label{experiments}
In this section, we will describe our experimental evaluations in details. Firstly, experiment settings including training details, evaluation metrics and datasets are introduced. Then, comparisons among the proposed SACCN with some popular state-of-the-art approaches are performed on several public benchmarks. Finally, in ablation studies, the effects of key components proposed in SACCN is investigated and analyzed.

\subsection{Experimental Setup}
\subsubsection{Training Details}
The network is trained end-to-end using the Adam optimizer with a learning rate of 0.0001 and a momentum of 0.9 on single NVIDIA 2080Ti GPU. The batch size is set to be 4. Mean Squared Error (MSE) is taken as the loss function to constrain the network output and ground truth. Following HA-CCN, we employ similar data augmentation methods on training data, such as randomly cropping $400\times400$ image patches from the images, and randomly flipping patches horizontally with a probability of 0.5.

\subsubsection{Evaluation Metrics}
Following most of crowd counting methods, we use mean absolute error (MAE) and mean square error (MSE) as evaluation metrics. The smaller the value of MAE and MSE, the better the performance.
\begin{equation}
\small
	MAE = \frac{1}{N} \sum _{i=1}^N\vert C_i-\hat{C_i}\vert
\end{equation}
\begin{equation}
\small
	MSE = \sqrt{\frac{1}{N} \sum _{i=1}^N\vert C_i-\hat{C_i}\vert^2}
\end{equation}
where, $C_i$ is the predicted counting number, and $\hat{C_i}$ stands for the ground truth. 

For vehicle counting, we use GAME metric\cite{guerrero2015extremely} on TRANCOS, as defined in Equation~\ref{gamemetric}:
\begin{equation}
\label{gamemetric}
\small
	\text{GAME}^L = \frac{1}{N} \sum _{i} \sum _{l=1}^{2^L}\vert C_i^l-\hat{C_i^l}\vert
\end{equation}
where $\text{GAME}^L$ means the case that images are divided into $2^L$ non-overlapping regions.

\subsection{Datasets}
We conduct experiments on several benchmark datsets, including ShanghaiTech Part\_A and Part\_B~\cite{zhang2015cross}, UCF\_CC\_50~\cite{idrees2013multi}, UCF-QNRF~\cite{idrees2018composition}, WorldExpo'10~\cite{zhang2015cross}, Mall~\cite{chen2012feature}. and the TRANCOS~\cite{guerrero2015extremely} dataset.

\subsubsection{\textbf{ShanghaiTech Dataset}} 
ShanghaiTech dataset~\cite{zhang2015cross} is composed of 1198 annotated images, including 330165 annotated people. The dataset consists of Part\_A and Part\_B. Part\_A contains 482 images of highly congested scenes randomly downloaded from the Internet. Among them, 300 images are for training and 182 images are for the test. Part\_B contains 716 images of relatively sparse crowd scene taken from streets in Shanghai. There are 400 images in the training set and 316 images in the test set.

\begin{table}
	\centering
	\setlength{\tabcolsep}{2.4mm}
		\caption{Comparison results of different methods on the WorldExpo'10 Dataset. Best results and Second results are highlighted.}
	\begin{tabular}{|l|c|c|c|c|c|c|}
		\hline
		Method & S1 & S2 & S3 & S4 & S5 & Average\\
		\hline
		MCNN\cite{zhang2016single}  & 3.4 & 20.6 & 12.9 & 13.0 & 8.1 & 11.6\\
        MSCNN\cite{zeng2017multi} & 7.8 & 15.4 & 14.9 & 11.8 & 5.8 & 11.7\\
        CP-CNN\cite{sindagi2017generating} & 2.9 & 14.7 & 10.5 & 10.4 & 5.8 & 8.86\\
        ic-CNN\cite{ranjan2018iterative} & 17.0 & 12.3 & 9.2 & 8.1 &  4.7 & 10.3\\
        SANet\cite{cao2018scale} & 2.6 & 13.2 & 9.0 & 13.3 & 3.0 & 8.2\\
		CSRNet\cite{li2018csrnet} & 2.9 & 11.5 & 8.6 & 16.6 & 3.4 & 8.6\\
		ADCrowdNet\cite{liu2019adcrowdnet} & 1.7 & 14.4 & 11.5 & \textbf{7.9} & 3.0 & 7.7\\
        CAN\cite{liu2019context} & 2.9 & 12.0 & 10.0 & \textbf{7.9} & 4.3 & 7.4\\
        MACC\cite{jiang2019mask} & 2.2  & 11.5 & 11.6 & 13.9 & 2.5 & 8.34\\
        COBC\cite{liu2019counting} & 2.1 & 11.5 & \textbf{8.1} & 16.8 & 2.5 & 8.2\\
		DSSINet\cite{liu2019crowd} & \textbf{1.6} & 9.5 & 9.5 & 10.4 & \textbf{2.5} & \textbf{6.7}\\
		ZoomCount\cite{sajid2020zoomcount} & 2.1 & 15.3 & 9.0 & 10.3 & 4.5 & 8.3\\
		SDANet\cite{miao2020shallow} & 2.0 & 14.3 & 12.5 & 9.5 & \textbf{2.5} & 8.1\\
		\hline
		SACCN (Ours) &\textbf{1.5}&\textbf{10.6}&\textbf{8.1}&8.9& 2.7 &\textbf{6.4}\\
		\hline	
	\end{tabular}
	\label{world}
\end{table}

\subsubsection{\textbf{UCF\_CC\_50 Dataset}} 
UCF\_CC\_50 dataset~\cite{idrees2013multi} is an extremely challenging crowd dataset. All 50 images of different resolutions are randomly collected from the Internet. The number of people in these images varies greatly in a wide range from 94 to 4543. There are a total of 63974 head annotations. The average number per image is 1280. Besides, the dataset contains diverse scenes with varying perspective distortions. For better verification of model accuracy, we follow standard setting~\cite{idrees2013multi}  and conduct 5-fold cross-validation for the test.

\subsubsection{\textbf{UCF\_QNRF Dataset}} 
UCF\_QNRF dataset~\cite{idrees2018composition} is a recently proposed large-scale crowd dataset. It contains 1,535 images in which there are about 1.25 million head annotations. These images have many different scenes, backgrounds, and perspectives. The training dataset contains 1201 training images. The test dataset containing 334 images. Many of them are in very high-resolution. Therefore, we transform all images to a fixed size of $1024\times768$ before any pre-processing. 

\subsubsection{\textbf{WorldExpo'10 Dataset}} 
WorldExpo'10 dataset~\cite{zhang2015cross} is a large scale data-driven cross-scene crowd counting dataset which is collected from Shanghai 2010 WorldExpo. The dataset consists of 1132 annotated video sequences captured by 108 different surveillance cameras. Each camera with different view angles provides the region of interest (ROI). The whole dataset contains a total of 3920 frames in a fixed resolution of $576\times720$. It is divided into a training set with 3380 frames and a test set with 600 frames. 

\subsubsection{\textbf{Mall Dataset}} 
Mall dataset~\cite{chen2012feature} is a dataset with relatively sparse crowd scenes. The dataset contains 2000 frames in a video, which are captured by using a surveillance camera installed in a shopping mall. The scenes significantly vary in the crowd’s scale and appearances due to perspective distortion. In addition, there exist challenging occlusions caused by objects in the scene. We use the first 800 frames as the training set, and the left 1200 frames as the test set to evaluate our model.

\begin{figure*}[!t]
	\centering
	\includegraphics[scale=0.4]{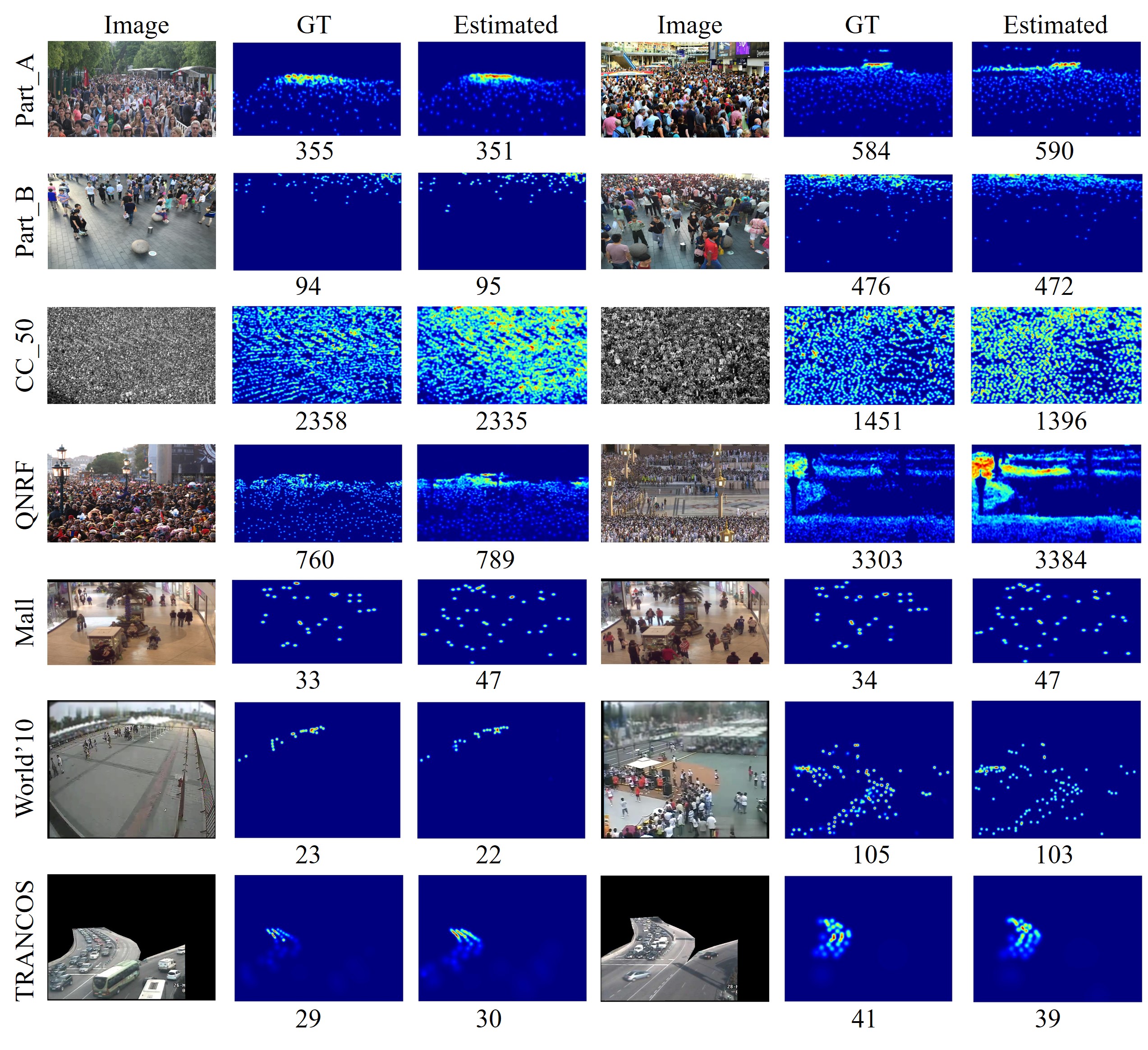} 
	\caption{The visualization of experiment samples on different datasets. The number below image indicates the number of crowd.}
	\label{result_data}
\end{figure*}

\begin{figure*}[h]
	\centering
	\includegraphics[scale=0.6]{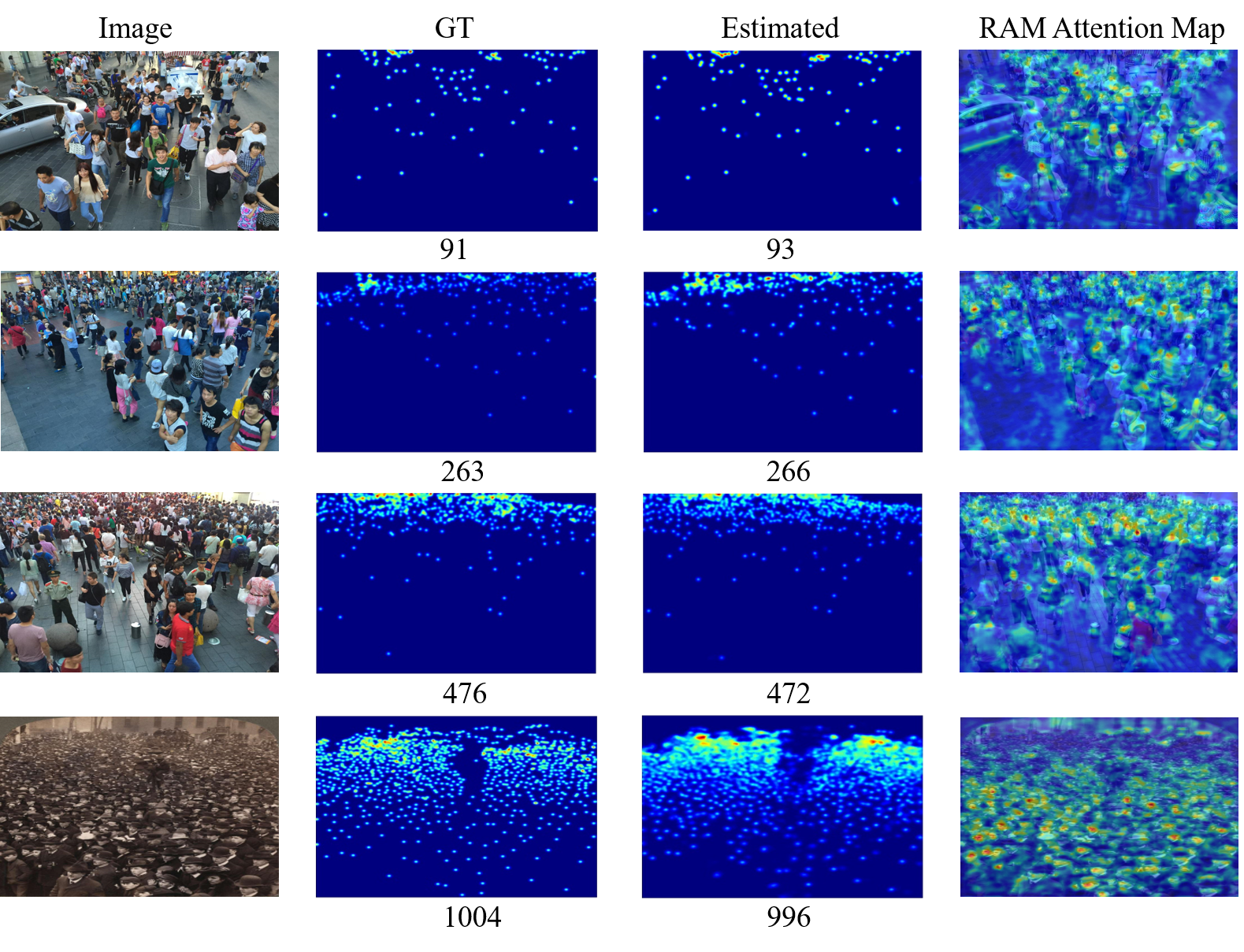} 
	\caption{Visualization of estimated density map and intermediate RAM attention map of SACCN. The number below image indicates the number of crowd.}
	\label{result}
\end{figure*}

\begin{table}
	\centering
	\setlength{\tabcolsep}{2mm}
		\caption{Comparison results of different methods on the TRANCOS Dataset. Best results and Second results are highlighted.}
	\begin{tabular}{|l|c|c|c|c|}
		\hline
		Method & GAME 0 & GAME 1 & GAME 2 & GAME 3 \\
		\hline
		Fiaschi et al.\cite{fiaschi2012learning}     & 17.77 & 20.14 & 23.65 & 25.99\\
        Lempitsky et al.\cite{lempitsky2010learning} & 13.76 & 16.72 & 20.72 & 24.367\\
        Hydra-3s\cite{onoro2016towards}              & 10.99 & 13.75 & 16.69 & 19.32\\
		CSRNet\cite{li2018csrnet}                    & 3.67 & 5.82 & 8.53 & 14.40\\
		PSDDN\cite{liu2019point}                     & 4.79 & 5.43 & 6.68 & \textbf{8.40}\\
		COBC\cite{liu2019counting}                   & 3.15 & 5.45 & 8.34 & 15.02\\
		HSRNet\cite{zou2020crowd}                    & \textbf{3.03} & \textbf{4.57} & 6.46 & 9.68\\
		KDMG\cite{wan2020kernel}                     & 3.13 & 4.79 & \textbf{6.20} & 8.68\\
		\hline
		SACCN (Ours)                                         &\textbf{2.20}&\textbf{3.11}&\textbf{4.18}&\textbf{5.77}\\
		\hline	
	\end{tabular}
	\label{car}
\end{table}

\begin{table}
	\centering
	\setlength{\tabcolsep}{8.5mm}
		\caption{Comparison results of different methods on the Mall Dataset. Best results and Second results are highlighted.}
	\begin{tabular}{|c|c|c|}
		\hline
		Method & MAE & MSE \\
		\hline
		CNN-Boosting\cite{walach2016learning} & 2.01&-\\
		W-VLAD\cite{sheng2016crowd} & 2.67 & 11.72\\
		ConvLSTM\cite{xiong2017spatiotemporal} & 2.24&8.5\\
		DecideNet\cite{liu2018decidenet} &1.52&1.9\\
		E3D\cite{zou2019enhanced} &1.64&2.13\\
		SAAN\cite{hossain2019crowd} & \textbf{1.28}& \textbf{1.68} \\
		AT-CNN\cite{zhao2019leveraging} & 2.28 & 2.9\\
		Deem-CFCN\cite{zhao2019scale} & 2.11 & 2.71\\
		FMLF\cite{ding2020crowd} &1.85&2.34\\
		HSRNet\cite{zou2020crowd} &1.8&2.28\\
		\hline
		SACCN (Ours) &\textbf{1.42}&\textbf{1.85}\\
		\hline	
	\end{tabular}
	\label{mall}
\end{table}

\subsubsection{\textbf{TRANCOS Dataset}} 
TRANCOS dataset~\cite{guerrero2015extremely} is the first one for vehicle counting in traffic jam scenes. It contains 1244 traffic images with vehicle numbers varying from 9 to 107. The dataset is often used to evaluate the generalization ability of population counting methods.

\subsection{Experiment Results}
We compare our proposed SACCN with state-of-the-art methods on ShanghaiTech Part\_A/Part\_B, UCF\_CC\_50, UCF\_QNRF, WorldExpo'10, Mall, and TRANCOS. 

From TABLE~\ref{table1}, on the ShanghaiTech Part\_A/Part\_B, UCF\_CC\_50, and UCF\_QNRF, it is not difficult to observe that the proposed SACCN achieves the lowest MAE in almost all datasets, showing more superiority over the other methods. 

The comparison results on the WorldExpo'10 dataset are shown in TABLE~\ref{world}. From TABLE~\ref{world}, we can see that the proposed SACCN delivers top-2 accuracy in 3 out of 5 scenes and achieves the best accuracy on average. 

For the Mall dataset and TRANCOS dataset, the comparison results are shown in TABLE~\ref{mall} and TABLE~\ref{car} respectively. In the case of the Mall dataset, the proposed SACCN achieves second lower MAE and MSE among all methods. In the case of the TRANCOS dataset, the proposed method delivers the best results according to the GAME metric. 

All the quantitative comparison results on different datasets indicate that our method is extremely robust to changes in scale, perspective and density. More visualized results on different datasets are shown in Fig.~\ref{result_data}. The counting numbers are recorded below each GT and estimated density map.

From Fig.~\ref{result_data}, it can be seen that the distribution of the generated density map by SACCN is very similar to the distribution of GT. Even though with a large span of counting numbers in Fig.~\ref{result_data}, the proposed SACCN can still accurately estimate the density map, which implies its robustness. 

In addition, in Fig.~\ref{result}, to display the effect of RAM, we visualize the RAM attention map for some samples with different backgrounds, perspectives, densities, and crowd scales. From Fig.~\ref{result}, we can observe that the RAM attention map accurately highlights the crowd areas and suppresses the non-crowd areas.
This shows that RAM components can effectively distinguish the crowd, which helps to count more accurately.

\subsection{Ablation Studies}
In this section, we conduct several ablation studies to demonstrate the effects of key components used in the proposed SACCN. Moreover, we further verify the influences of the proposed components on mitigating cluttered background and scale variation. All ablation experiments are conducted on ShanghaiTech Part\_B dataset.

\begin{table}   
	\centering
	\setlength{\tabcolsep}{2mm}
	\caption{The effects of the proposed components on ShanghaiTech Part\_B. DC represents the regional attention based dense connections. SC is the regional attention based skip-connections.}
	\begin{tabular}{|l|c | c |ccc cc|}
		\hline
		Method &MAE & MSE & RAM & SAM & AMM & DC & SC \\
		\hline
		SACCN-ram & 8.3 & 12.9 & - & $\checkmark$ & $\checkmark$ & $\checkmark$ & $\checkmark$ \\
		SACCN-sam & 7.2 & 11.7 & $\checkmark$ & - & $\checkmark$ & $\checkmark$ & $\checkmark$ \\
		SACCN-amm & 6.9 & 10.8 & $\checkmark$ & $\checkmark$ & - & $\checkmark$ & $\checkmark$ \\
		SACCN-dc & 7.0&11.0 & $\checkmark$ & $\checkmark$ & $\checkmark$ & - & $\checkmark$ \\
		SACCN-sc & 7.1 & 11.4 & $\checkmark$ & $\checkmark$ & $\checkmark$ & $\checkmark$ & - \\
		\hline
		SACCN & \textbf{6.8} & \textbf{10.5} & $\checkmark$ & $\checkmark$ & $\checkmark$ & $\checkmark$ & $\checkmark$ \\
		\hline		
	\end{tabular}
	\label{components}
\end{table}

\begin{table}   
	\centering
	\setlength{\tabcolsep}{4mm}
	\caption{The effects of different attention on ShanghaiTech Part\_B. CA is the channel attention and SP is the spatial attention.}
	\begin{tabular}{|l|c|c|ccc|}
		\hline
		Method &MAE & MSE  & CA & SP & RAM\\
		\hline
		$\text{SACCN}_{CA}$ & 7.8 & 12.3 & $\checkmark$ & - & -  \\
		$\text{SACCN}_{SP}$ & 7.6 & 11.5  & - & $\checkmark$ & - \\
		SACCN & \textbf{6.8} & \textbf{10.5} & - & - & $\checkmark$ \\
		\hline		
	\end{tabular}
	\label{attention}
\end{table}

\begin{figure*}[h]
	\centering
	\includegraphics[scale=0.5]{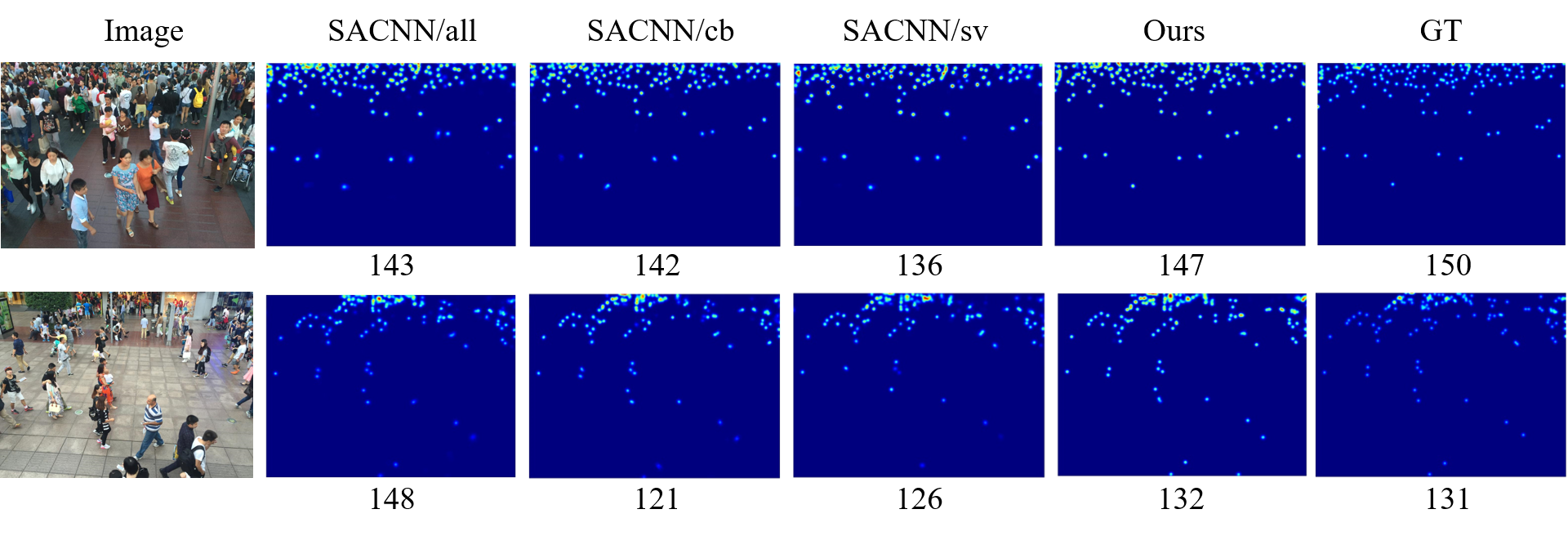} 
	\caption{Visualization on the effects of the proposed key components. The number below image indicates the GT or estimated number of crowd.}
	\label{rebuttal}
\end{figure*}
\subsubsection{\textbf{The effects of the proposed components}}
TABLE~\ref{components} shows the performance of all the proposed components used in the proposed SACCN. The \textit{DC} specifically represents the regional attention based dense connections. The \textit{SC} represents the regional attention based skip-connections. 

Firstly, without RAM, from the comparison between the first row "SACCN-ram" and the last row "SACCN", we can find that the performance has severely deteriorated, indicating that region identification is very helpful for crowd counting. 

Secondly, from the comparison between the second row "SACCN-sam" and the last row "SACCN", we can infer that at deep layers, using semantic self-attentions is beneficial. 
This is because the feature maps in deeper layers have stronger semantics and can help to distinguish the crowd and the background in semantic space.

Thirdly, to demonstrate the effect of asymmetric convolution, the ablation method "SACCN-amm" uses the convention convolution with $3 \times 3$ instead of the asymmetric convolution with $1 \times 3$ and $3 \times 1$, and use the convention convolution with $5 \times 5$ instead of the asymmetric convolution with $1 \times 5$ and $5 \times 1$ in AMM. From the comparison result in TABLE~\ref{components}, it is unsurprising to find the performance gap between convention convolution and asymmetric convolution is very small. However, we should point out that herein asymmetric convolutions have fewer parameters than the convention convolution, which can greatly reduce the model size.

Finally, after comparing the cases that remove dense connections or skip-connections from the proposed SACCN, we can conclude that combining features of different scales can improve the accuracy of crowd counting. This method can also prevent gradient vanishing caused by network deepening.

\subsubsection{\textbf{The effects of different attentions}}
In order to verify the effect of different attention, we respectively replace RAM with single channel attention(CA) and spatial attention (SP). 

In the TABLE~\ref{attention}, the first row shows the results that use CA instead of RAM. The second row shows the results that use SP instead of RAM. Through the performance comparison, we can see that the performance of only using CA or SP is worse than using RAM. Therefore, we have reasons to infer that, when distinguishing the crowd area from the background, both spatial attention and channel attention are important and complementary. 

\begin{table}   
	\centering
	\setlength{\tabcolsep}{2mm}
	\caption{The effects of the proposed components on mitigating cluttered background and scale variation.}
	\begin{tabular}{|l|cc|c|c|c|c|c|}
		\hline
		\multirow{2}{*}{Method} & \multirow{2}{*}{MAE} & \multirow{2}{*}{MSE} & \multicolumn{2}{|c|}{clutter background} & \multicolumn{3}{|c|}{scale variation}\\
		\cline{4-8}
		 &  &  & RAM & SAM & AMM & DC & SC \\
		\hline
		SACCN/all & 8.9 & 14.4 & - & - & - & - & - \\
		SACCN/cb & 8.3 & 13.7 & - & - & $\checkmark$ & $\checkmark$ & $\checkmark$ \\
		SACCN/sv & 7.7 & 12.7 & $\checkmark$ & $\checkmark$ & - & - & - \\
		SACCN & 6.8 & 10.5 & $\checkmark$ & $\checkmark$ & $\checkmark$ & $\checkmark$ & $\checkmark$ \\
		\hline		
	\end{tabular}
	\label{mitigate}
\end{table}
\subsubsection{\textbf{The effects of the proposed key components on mitigating adverse influences of cluttered background and scale variation}}
As we have summarized, cluttered background and scale variation are the main concentrated challenges in crowd counting. In this paper, we have proposed SACCN with some key innovative components to mitigate them. Among these components, \textit{RAM} and \textit{SAM} are attention modules mainly focusing on crowd identification from cluttered backgrounds. \textit{AMM}, \textit{DC} and \textit{SC} mainly focus on scale variation.
For a better understanding of the effects of these key components, we conduct more experiments (TABLE~\ref{mitigate}) to demonstrate their respective performance on mitigating the mentioned obstacles. The ablation results are shown in TABLE~\ref{mitigate}.

From TABLE~\ref{mitigate}, the comparison results indicate that each group of the proposed components can effectively improve the performance. Moreover, from Fig.~\ref{rebuttal}, we can observe that the density maps generated by the proposed SACCN are more similar to the GT. The method without the proposed components cannot accurately count the crowd in the image.

\section{Conclusion}
\label{conclusion}
In this paper, we proposed a Scale-Aware Crowd Counting Network (SACCN) that can efficiently deal with the task of crowd counting under clutter background and scale variations. In the proposed SACCN, regional attention and semantic attention are employed for highlighting crowd area. Meanwhile, the asymmetric multi-scale module is specifically designed to handle the scale variations problem, and regional attention based connections are constructed for feature reuse and representation enhancement. Extensive experiments demonstrated that the proposed SACCN achieves superior performance against other state-of-the-art methods.


\ifCLASSOPTIONcaptionsoff
  \newpage
\fi



%
\bibliographystyle{IEEEtran}
\bibliography{ref_cc}

%




\end{document}